\documentclass{article}

\usepackage[final]{nips_2016}
\usepackage[utf8]{inputenc} % allow utf-8 input
\usepackage[T1]{fontenc}    % use 8-bit T1 fonts
\usepackage{hyperref}       % hyperlinks
\usepackage{url}            % simple URL typesetting
\usepackage{booktabs}       % professional-quality tables
\usepackage{amsfonts}       % blackboard math symbols
\usepackage{nicefrac}       % compact symbols for 1/2, etc.
\usepackage{microtype}      % microtypography

\usepackage{amsmath,amscd,amssymb}
\usepackage{graphicx}
\usepackage{algorithm}
\usepackage[noend]{algpseudocode}

\newcommand{\deldel}[2]{\frac{\partial #1}{\partial #2}}

\title{Adversarial Message Passing For Graphical Models}

\author{
Theofanis Karaletsos\\Geometric Intelligence\\theo@geometric.ai\\[2ex]
}

\begin{document}

\maketitle

\begin{abstract}
Bayesian inference on structured models typically relies on the ability to infer posterior distributions of underlying hidden variables.  However, inference in implicit models or complex posterior distributions is hard.
A popular tool for learning implicit models are generative adversarial networks (GANs) which learn parameters of generators by fooling discriminators.
Typically, GANs are considered to be models themselves and are not understood in the context of inference. Current techniques rely on inefficient global discrimination of joint distributions to perform learning, or only consider discriminating a single output variable.
We overcome these limitations by treating GANs as a basis for likelihood-free inference in generative models and generalize them to Bayesian posterior inference over factor graphs.
We propose local learning rules based on message passing minimizing a global divergence criterion involving cooperating local adversaries used to sidestep explicit likelihood evaluations. This allows us to compose models and yields a unified inference and learning framework for adversarial learning. Our framework treats model specification and inference separately and facilitates richly structured models within the family of Directed Acyclic Graphs, including components such as intractable likelihoods, non-differentiable models, simulators and generally cumbersome models.
A key result of our treatment is the insight that Bayesian inference on structured models can be performed only with sampling and discrimination when using nonparametric variational families, without access to explicit distributions. As a side-result, we discuss the link to likelihood maximization.
These approaches hold promise to be useful in the toolbox of probabilistic modelers and enrich the gamut of current probabilistic programming applications.
\end{abstract}

\section{\bf Introduction \& Related Work}
The typical setup of the modeler using variational techniques is to formulate a model hypothesis, choose an approximate variational model for inference with an appropriate variational family, combine those two models with a distance measure such as a divergence  that is appropriate for the inferential task and then match the statistics of these distributions given these constraints.
We propose to add an extra step of using adversaries to reparametrize relationships between distributions which are intractbale or inaccessible.
This leads to a treatment of adversarial learning from the perspective of distributed Bayesian inference on generative models, in particular variational inference. The task of Bayesian inference is to infer posterior distributions for all unobserved variables in a joint model, corresponding to marginal likelihood maximiziation of the model given evidence. 

We achieve generalization of adversarial learning to arbitrary structured models by introducing a local message passing algorithm based on adversaries and show that is is performing a clean approximation to a posterior defined by an explicit model.
We thus present novel work that explains and clarifies the separation of modeling and inference in the context of adversarial learning and opens the door to building flexible probabilistic programs using adversarial inference as a unified framework for inference, learning and generation given assumptions about the model.
In addition, our framework automatically leads to distributed adversarial inference with cooperating adversaries and clarifies how adversarial inference performs implicit likelihood maximization.

In recent work it has been shown that neural networks can be used as samplers for divergence minimization in a general class of divergences~\citep{nowozin2016f}. Furthermore, it was clarified in concurrent work very much in the same spirit with our paper such as ~\citep{uehara2016generative} and ~\citep{mohamed2016learning} that Generative Adversarial Networks can be seen as a form of inference on ratios of partition functions, with early links towards training generative models. First steps towards GANs on structured models were taken in recent papers like the SeqGAN~\citep{yu2016seqgan}, Professor Forcing~\citep{professorForcing} and~\citep{liu2016coupled}.
We highlight that a side-result of~\citep{sonderby2016amortised} is a derivation of a KL-divergence loss for standard GANs and the introduction of instance noise, both of which are related to results we discuss in our Appendix.
Finally, inference for a narrow class of specific fixed instances of models was introduced in similar fashion in~\citep{donahue2016adversarial},~\citep{dumoulin2016adversarially}  and ~\citep{makhzani2015adversarial}) using global adversaries, but not generalized to more flexible models.

\section{Generative Adversarial Networks}
\label{gan}
Basic GANs have been postulated to follow a value function playing an adversarial game between a discriminator $D$ with parameters $\xi$ and a generator $G$ with parameters $\theta$.
\begin{equation}
\begin{split}
\min\limits_{\theta} \max\limits_{\xi} V(\xi, \theta) &= \mathbb{E}_{x\sim p^{*}(x)} \text{log} D(x;\xi) + \mathbb{E}_{x \sim Q(x)}\text{log} (1-D(x;\xi))  \\
&= \mathbb{E}_{x\sim p^{*}(x)} \text{log} D(x;\xi) + \mathbb{E}_{z\sim P(z)}\text{log} (1-D(G(x;\theta);\xi))  \\
\end{split}
\end{equation}

For $m(x) = \frac{1}{2} p(x) + \frac{1}{2} q(x)$ an analogy can be shown between the value function and the following probabilistic formulation. 
\begin{equation}
\begin{split}
\text{\bf{JSD}}(q(x)||p(x))&= \frac{1}{2} \int \limits_{x^{*}} q(x^{*}) \text{log} \frac{q(x^{*})}{m(x)} dx +  \frac{1}{2}  \int \limits_{x^{*}} p(x^{*}) \text{log} \frac{p(x^{*})}{m(x)} dx\\
& = \frac{1}{2} \int \limits_{x^{*}} q(x^{*}) \text{log} \frac{q(x^{*})}{m(x)} dx +\frac{1}{2} \int \limits_{z} p(z) \text{log} \frac{p(x|z)}{m(x)} dz
\end{split}
\end{equation}

%\subsection{Graphical Model Inference}
%explain markov blanket\\
%d-separation\\
%parents, children of a variable\\
%... as such, each variable is fully specified by the states of its parents and children and the factors connecting them

\section{Approximate Inference in Graphical Models through Adversarial Learning}

We show, that instead of one large GAN discriminating between the joint distribution of all variables in graphical models (as done in \citep{donahue2016adversarial},~\citep{dumoulin2016adversarially} and \citep{makhzani2015adversarial}), we can perform distributed adversarial inference by discriminating locally for each variable whether it is a valid sample or not.
We can maximize these local discriminators to yield a globally convergent distributed learning procedure, {\it \bf adversarial message passing}.

We are given a joint distribution over $I$-many variables $p({\bf X}) = p(x_{0}, ..,x_{I})$ with a graph structure $\mathcal{G}$ and a factorization given by the computational graph $p({\bf X}) = \prod \limits_{i}p(x_i|\text{pa}(x_i))$, where $\text{pa}(x_{i})$ denote the parents of variable $x_{i}$ in $\mathcal{G}$. We can derive an inverse factorization  $q({\bf X}) = \prod \limits_{i}q(x_{i}|\tilde{\text{pa}}(x_{i}))$ which preserves the variable dependence structure. In the inverse factorization, we consider $\tilde{ \text{pa}}(x_{i})$ to be the part of the Markov blanket for the variable $x_i$ needed in order to {\it d-seperate} it given observations.
These factorizations have been explained at length in the context of stochastic inversion~\citep{stuhlmuller2013learning} and form a structured inverse factorization as used in variational inference~\citep{hoffman2015structured}, while also being widely used in the message passing literature~\citep{winn2005variational},\citep{minka2001expectation}.

%We identify two learning settings. First, if a specified, potentially intractable, graphical model exists and we want to learn to infer data.
%Second, if we are given a dataset ${\bf x_1}$ and a model structure $\mathcal{M}$, we may want to fit that model, meaning that the factors for the model need to be learned.

\begin{equation}
P({\bf X}) = P(x_1, x_2, ..., x_D) = \prod \limits_{i=1}^{D}P(x_i| \text{pa}(x_i)).
\end{equation}

We use factorizations of dependencies as the basis to derive schemes for Bayesian Learning and inference which take advantage of adversarial learning.

\subsection{Adversarial Message Passing For JS-Divergence Minimization}

In this section, we match the local Jensen-Shannon divergence ({\bf JSD}) of variables to perform approximate inference locally.

We use the intuition that we wish to match the local statistics of approximations to the posterior by minimizing a divergence $\text{Div}$ at each factor indexed by $i$, $\text{Div}\Big( q(x^{*}_i,\tilde{\text{pa}}(x_i)) ||p(x_i,\text{pa}(x_i)) \Big)$. This is a typical assumption in divergence based message passing~\citep{minka2005divergence}.\\

Given a definition of $m(x_i , \text{pa}(x_{i})) = \Big[0.5 q(x_i , \tilde{\text{pa}}(x_{i}))+0.5 p(x_i , \text{pa}(x_{i}))\Big]$, we can express local minimization of the {\bf JSD} as a sum of divergences, compactly written as follows:
\begin{equation}
\begin{split}
\text{Div}_{loc}\Big( q({\bf X}) ||p({\bf X} )  \Big) = & \frac{1}{2}\int \limits_{x_{0}} p^{*}(x_{0}) ... \int \limits_{x_{I}} q(x_{I}|\tilde{\text{pa}}(x_{I})) \text{log} \frac{\prod \limits_{i=1}^{I}q(x_{i-1} , \tilde{\text{pa}}(x_{i-1}))}{\prod \limits_{i=1}^{I} m(x_{i-1} , \text{pa}(x_{i-1})) } dx_{0...I}\\
&+ \frac{1}{2} \int \limits_{x_{I}} p(x_{I}) ... \int \limits_{x_{0}} p(x_{0}|\text{pa}(x_{0})) \text{log} \frac{\prod \limits_{i=0}^{I-1}p(x_i , \text{pa}(x_{i}))}{\prod \limits_{i=0}^{I-1} m(x_i , \text{pa}(x_{i}))} dx_{0...I}
\end{split}
\label{eq:local_jsd}
\end{equation}

We rephrase the above divergence in terms of a sum of the local adversaries by noting that each factor can be expressed as an expectation over the score of the class the discriminator will assign to the bottom-up and top-down samples.

We can use an optimal discriminator $D_{i}^{*}$ as an adversary at each local factor $i$ to express ratios of distributions $D_{i}^{*}(x_i , \text{pa}(x_{i})) = \frac{ p(x_i , \text{pa}(x_{i}))}{m(x_i , \text{pa}(x_{i}))}$ and $1-D_{i}^{*}(x_i , \text{pa}(x_{i})) = \frac{ q(x_i , \tilde{\text{pa}}(x_{i}))}{m(x_i , \text{pa}(x_{i}))}$.
In order to calibrate these adversaries, we can derive a loss function $\mathcal{L}_{locD}$ and train models to discriminate between inference and model samples generated during training.

Combining these adversaries with Equation~\ref{eq:local_jsd} yields a reparametrized form of the divergence term: 
\begin{equation}
\begin{split}
\text{Div}_{loc}\Big( q({\bf X}) ||p({\bf X} )  \Big) = & \frac{1}{2}\int \limits_{x_{0}} p^{*}(x_{0}) ... \int \limits_{x_{I}} q(x_{I}|\tilde{\text{pa}}(x_{I})) \text{log}\Big[ \prod \limits_{i=1}^{I} \Big( 1-D^{*}_{i}(x_{i-1},\tilde{\text{pa}}(x_{i-1})) \Big) \Big] dx_{0...I}  \\
&+ \frac{1}{2} \int \limits_{x_{I}} p(x_{I}) ... \int \limits_{x_{0}} p(x_{0}|\text{pa}(x_{0})) \text{log} \Big[ \prod \limits_{i=0}^{I-1} \Big( D^{*}_{i}(x_{i} , \text{pa}(x_{i})) \Big) \Big] dx_{0...I}
\end{split}
\end{equation}

This joint term can be approximated efficiently across each local term by performing bottom-up sampling of $L$ particles through inference models and $K$ top down samples from the prior. This procedure yields two Markov chains transitioning from evidence to prior and from prior to evidence in a setting similar to that used for the Bennett acceptance ratio estimator~\citep{bennett1976efficient} and related newer work~\citep{geyer1991reweighting,shirts2008statistically,liu2015estimating,carlson2016partition,grosse2015sandwiching}.

We consider generative models to be parameterized by parameters $\theta$ capturing the generative factors and inverse models performing inference over unobserved variables $X_u$ and observed variables $X_o$ to be parametrized by $\phi$ denoting variational parameters or parameters of inference models. Learned adversaries have parameters $\xi$. We obtain the following objective function for learning graphical models using the above:
\begin{equation}
\mathcal{L}_{locM}(\theta,\phi|{\bf X}) =  \text{Div}_{loc}\Big(q({\bf X}| \phi))||p({\bf X}|\theta)\Big)
\end{equation}
Concurrently, since the variable-wise adversaries $D_{i}(\cdot|\xi)$ need to be trained to approximate optimality, we can derive a loss function for them as follows:
\begin{equation}
\mathcal{L}_{locD}(\xi|{\bf X})= -\Big[ \mathbb{E}_{x_{i},\text{pa}(x_i)} \text{log}D_{i}(x_i,\text{pa}(x_i)) + \mathbb{E}_{x_{i-1},\tilde{\text{pa}}(x_{i-1}) } \text{log}(1-D_{i}(x_{i-1},\tilde{\text{pa}}(x_{i-1}) ))\Big]
\end{equation}

%It is easy to show that this introduced local divergence is related to the %global Jensen Shannon divergence ({\bf JSD}) in the following way:
%\begin{equation}
%\text{{\bf D}}_{loc}\Big( q({\bf X}) ||p({\bf X} )  \Big) \leq \text{JS}\Big( %q({\bf X}) ||p({\bf X} )  \Big)
%\end{equation}
Equality to the {\bf JSD} holds when for each factor $i$  we minimze the divergence between the approximation and the true distribution, obtaining $\tilde{\text{pa}}(x_{i})) = p(x_i | \text{pa}(x_{i}))$ This also reveals that the fixed points of $ \text{{\bf Div}}_{loc}$ are the fixed points of {\bf JSD}, which correspond to global fixed points to the true distribution.
In general, $\text{{\bf Div}}_{loc}$ provides a looser divergence than {\bf JSD}, which intuitively makes sense since it performs a local calculation through message passing and formally can be shown by comparing the denominators in the respective divergence terms.

\subsection{Distributed Adversarial Message Passing}
We obtain the following practical benefits through distribution of adversarial divergence calculations along a graph:
\begin{enumerate}
\item In the traditional adversarial framework, calculating the global {\bf JSD} requires learning and evaluation of a discriminator over the joint distribution. For larger graphical models with multiple potentially high-dimensional variables, this quickly becomes impossible or impractical.
\item As long as the adversary is far away from the Bayes-Optimal discriminator, the assumption to reparametrize the ratio-term through the discriminator is not fulfilled. Local discriminators have a better chance of obtaining locally strong solutions for smaller tuples of variables than global discriminators of an entire graphical model state.
\item Local discriminators furthermore permit interesting learning settings, like partial observability as occuring in semi-supervised learning, time-series with irregular time-steps, multi-modal data-sets with missing modalities and more.
\end{enumerate}

With our framework, we perform local discrimination per factor and achieve a similar computation to that of a global discriminator needed for the global {\bf JSD} or {\bf KLD} to hold, see Algorithm~\ref{alg:adv_mp}.

\begin{algorithm}[h]
\caption{Adversarial Message Passing}\label{alg:adv_mp}
\begin{algorithmic}[1]
\Procedure{ADMP}{$X,iter$}\Comment{X: a given dataset, iter: \# of iterations}
\State $\phi_0 \sim P(\phi_{\text{init}})$
\State $w_0 \sim P(w_{\text{init}})$ \Comment{initialize weights of prior and model approximation}
\State ${\bf \epsilon_0 } \sim p({\bf \epsilon})$ \Comment{Initial Noise-vector}
%\State $\phi \gets a\bmod b$
\For{$t\leq iter$}\Comment{Loop over iterations}
\For{$X_t \in X$} \Comment{Sample minibatch $X_t$}
\State $\forall i: x_{i}^{l} \sim q(x_i | \tilde{\text{pa}}(x_i))$ \Comment{Infer parents of each variable with inference model}
\State $\forall i: x_{i}^{k} \sim p(x_i | \text{pa}(x_i))$ \Comment{Sample from model (using $\theta$ or specified model)}
\State ${\bf \epsilon_t } \sim p({\bf \epsilon})$ \Comment{Sample an appropriate noise vector}
\For{ $i$ in factors}\Comment{Cycle through factors and update parameters}
\State $\xi_{t,i} \gets \xi_{t-1,i} -\deldel{\mathcal{L}_{locD}(\theta_{t-1},\phi_{t-1},\xi_{t-1}; \epsilon_t , X_t)}{\xi} $
\State $\theta_{t,i} \gets \theta_{t-1,i} - \deldel{\mathcal{L}_{locM}(\theta_{t-1},\phi_{t-1},\xi_{t-1}; \epsilon_t , X_t)}{\theta_{t-1}}$
\State $\phi_{t,i} \gets \phi_{t-1,i} -\deldel{\mathcal{L}_{locM}(\theta_{t-1},\phi_{t-1},\xi_{t-1}; \epsilon_t , X_t)}{\phi} $
\EndFor\label{factors}
\EndFor\label{minibatchFor}
\EndFor\label{advmpfor}
\State \textbf{return} $\theta_t, \phi_t, \xi_t$\Comment{Parameters for the adversaries $\xi$, variational approximations $\phi$, model $\theta$ learned from data X}
\EndProcedure
\end{algorithmic}
\end{algorithm}

\section{Discussion}
Adversarial Message Passing provides a framework to perform likelihood-free inference for graphical models. It furthermore enriches the family of message passing algorithms by a previously intractable divergence class and faciliates the usage of nonparametric variational families for learning and inference in likelihood-free and cumbersome graphical models. We note that more general classes of divergences such as f-divergences and the newly proposed $\chi$-divergence~\citep{dieng2016chi} fall under this framework when considering adversaries of different structure, since adversaries serve as function approximations to score relations of distributions and can be composed locally to infer larger models. In the appendix we exhibit similar treatments for KL-divergence as an example, yielding results that use only reconstructive sampling in a graph. Interestingly, this allows us to cleanly derive combinations of adversarial loss functions with explicit parametric losses mapping to likelihood maximization, as empirically used by various previous papers without formal justification.
It is also easy to mix different divergences locally depending on suitability. Furthermore, a generalization of the work presented here can use MMD~\citep{gretton2012kernel} to perform local approximations in computational graphs. 
Finally, we suggest that the introduced message passing scheme can be generalized to undirected graphical models.

\section*{Acknowledgements}
We thank Eli Bingham, John Chodera, Noah Goodman and Zoubin Ghahramani for helpful and inspiring discussions. We furthermore acknowledge Anh Nguyen and Jason Yosinski for empirical demonstrations of the benefits of adversarial learning.

\bibliographystyle{apalike}
\bibliography{ganNIPS_arxiv16}

\newpage
\section{Appendix}

\subsection{Learning Deep Generative Models}
We exemplify how to use the introduced framework at the example of a deep generative model with two stochastic layers, applied to modeling MNIST digits.

\begin{figure}[h]
\begin{center}
\includegraphics[width=1\linewidth]{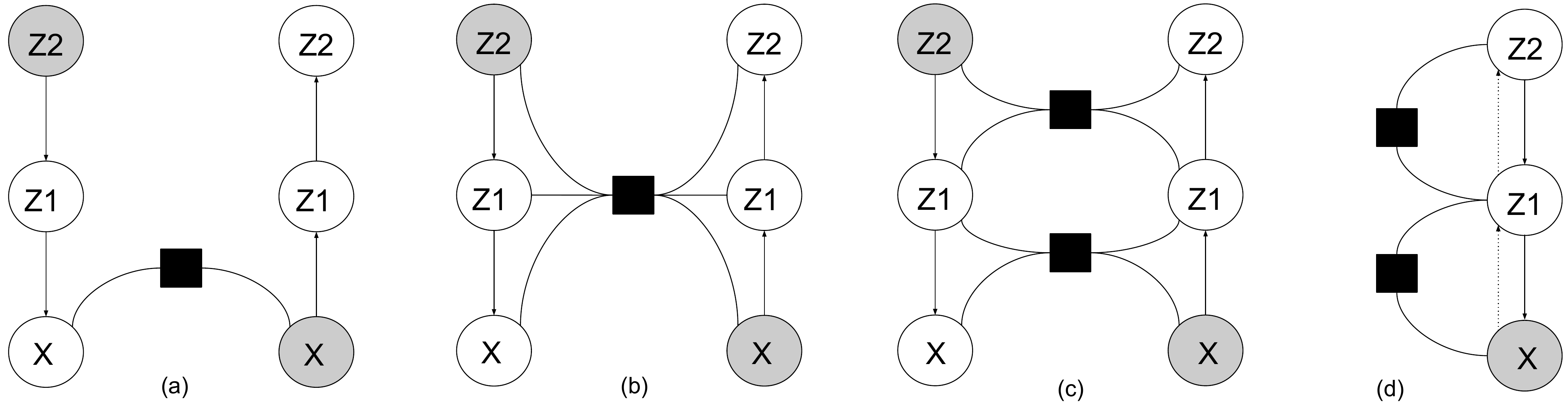}
\end{center}
\vspace{-5pt}
\caption{We show the four different learning variants. Black boxes indicate adversaries connected to their input variables.
%The models differ in how they encode semantics in the embeddings. 
({\bf a}) A standard adversarial network ({\bf GAN}) which only has to generate observable $X$ ({\bf b}) A deep variant of a global bidirectional adversarial network ({\bf BiGan,ALI}) ({\bf c}) A model using adversarial message passing with JSD minimization using local adversaries ({\bf ADMP-JSDloc}) ({\bf d}) A model using adversarial message passing with KL minimization using local adversaries ({\bf ADMP-KL}).}
\label{figADV}
% \vspace{-17pt}
\end{figure}

The generative model is defined as follows:
\begin{enumerate}
\item $z_2 \sim P(z_2)$
\item $z_1 \sim P(z_1|z_2)$
\item $x \sim P(x|z_1)$
\end{enumerate}

We use two adversaries $D_{1}(x,z_1)$ and $D_{2}(z_1,z_2)$ to drive learning. The inverse factorization here is trivial since Markov blankets on chain-graphs form unique tupels of variables.
We show the different inferential strategies in Figure~\ref{figADV}.

We note that compared to the usual application of GANs, we explicitly define the model here. For instance, $P(z_2)=\mathcal{N}(0,1)$, $P(z_1|z_2) = \mathcal{N}(\mu_{z_2},\Sigma_{z_2})$, $P(x|z_1) = Ber(\mu_{z_1})$.Interestingly, when we generate from the priors we also sample observation noise from the Bernoulli likelihood. This yields similar results to what is defined as instance noise in~\citep{sonderby2016amortised}, since a layer of noise is added to all generated images before they are passed into adversaries.

\subsection{Inverse Factorizations of Graphical Models}

\begin{figure}[h]
\begin{center}
\includegraphics[width=1\linewidth]{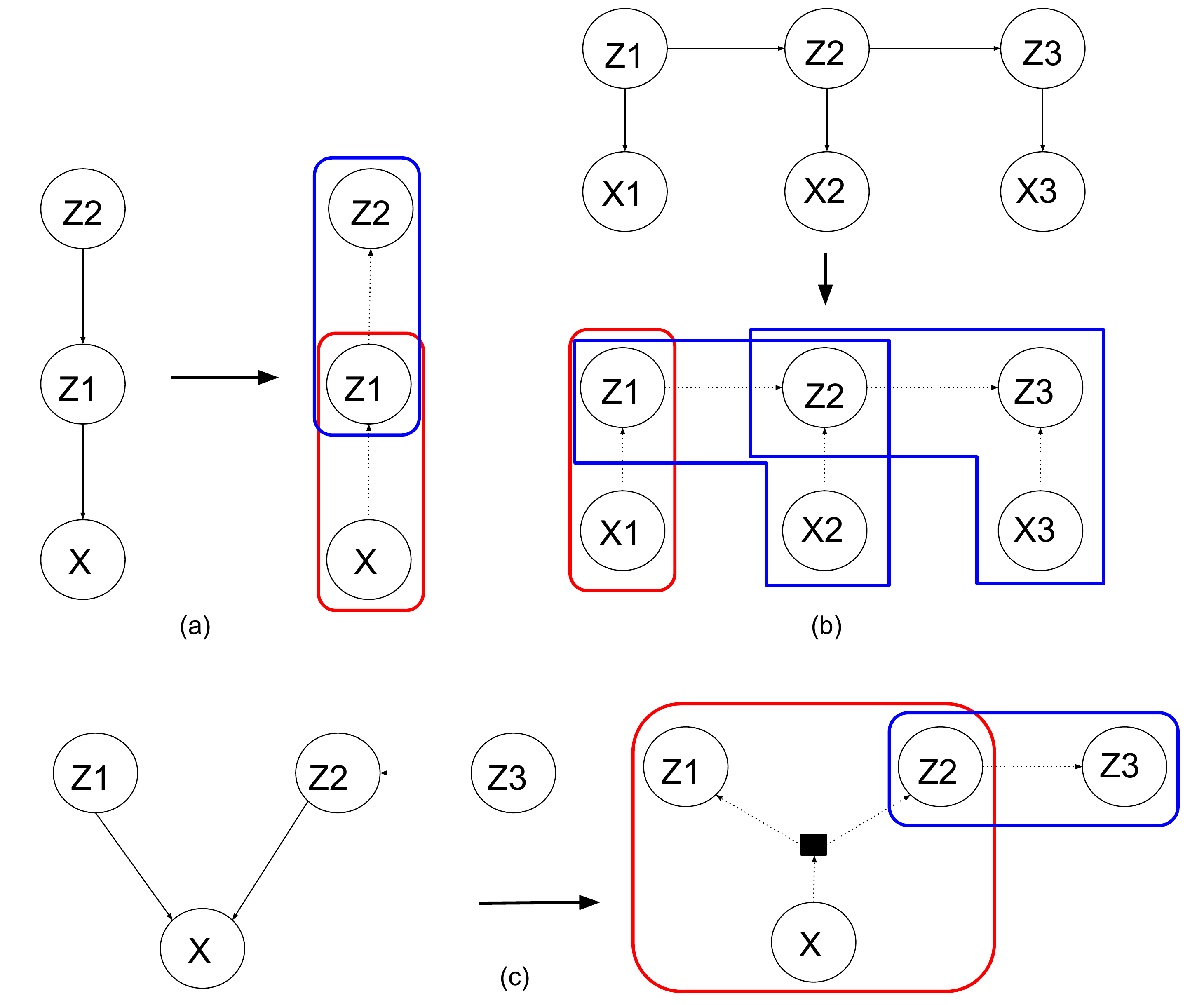}
\end{center}
\vspace{-5pt}
\caption{ We show three model variants and their inverse factorization into different cliques, here denoted by different colours. ({\bf a}) a two layer deep generative model  ({\bf b}) a state-space model over time  ({\bf c}) a multifactorial model }
\label{invFact}
% \vspace{-17pt}
\end{figure}

The proposed framework heavily relies on our ability to generate inverse factorization of graphical models. In Figure~\ref{invFact} we show how many of these factorizations arise naturally from the model structure, although it is not always a necessity that this is the optimal structure for an inverse model.

\subsection{Derivations for Variational Inference}
For a model $P(x,z)$ with variable $z$ we can derive the following:
\begin{equation}
\begin{split}
\text{KL}(q(z|x)||p(z|x))&=\int \limits_{z} q(z|x) \text{log} \frac{q(z|x)}{p(z|x)} dz \\
& = \int \limits_{z} q(z|x) \text{log} \frac{q(z|x)p(x)}{p(x,z)} dz \\
& = \int \limits_{z} q(z|x) \text{log} \frac{q(z|x)p(x)}{p(z)p(x|z)} dz \\
& = \int \limits_{z} q(z|x) \text{log} \frac{q(z|x)}{p(z)p(x|z)} dz + \text{log}p(x)\\
& = \int \limits_{z} q(z|x) \text{log} \frac{q(z|x)}{p(z)} dz - \int \limits_{z} q(z|x) \text{log}p(x|z)dz + \text{log}p(x)\\
\text{log}p(x) &= \int \limits_{z} q(z|x) \text{log}p(x|z)dz - \int \limits_{z} q(z|x) \text{log} \frac{q(z|x)}{p(z)} dz + \text{KL}(q(z|x)||p(z|x))\\
\text{log}p(x) & \geq \int \limits_{z} q(z|x) \text{log}p(x|z)dz - \int \limits_{z} q(z|x) \text{log} \frac{q(z|x)}{p(z)} dz\\
\text{log}p(x) & \geq \int \limits_{z} q(z|x) \text{log}p(x|z)dz - \text{KL}(q(z|x)||p(z))\\
\end{split}
\end{equation}

\subsection{Generative Adversarial Networks For KL-divergence minimization}
Assuming $D(x) = \frac{p(x)}{q(x)+p(x)}$ and $(1-D(x)) = \frac{q(x)}{q(x)+p(x)}$ and $D(x)$ being a Bayes-optimal discriminator  , we can derive the following divergence:
\begin{equation}
\begin{split}
\text{KL}(q(x)||p(x))&=\int \limits_{x} q(x) \text{log} \frac{q(x)}{p(x)} dx \\
& = \int \limits_{x} q(x) \text{log} \frac{1-D(x)}{D(x)} dx \\
\end{split}
\label{eq:gankl}
\end{equation}

This has also been considered as a loss function for adversarial learning in recent work on image super-resolution~\citep{sonderby2016amortised}.

%The value function for the adversarial game then is

%\begin{equation}
%\begin{split}
%V(\xi, \theta)= \max\limits_{\xi,\theta} \mathbb{E}_{x\sim p*(x)} \text{log} D(x;\xi) + \mathbb{E}_{z \sim P(z)} \Big[ \text{log}(D(p(x|z);\xi)) -\text{log}(1-D(p(x|z);\xi))\Big]
%\end{split}
%\end{equation}

\subsection{Adversarial Message Passing For KL-Divergence Minimization}
In the following we will derive two distinct learning rules which will enable us to perform implicit divergence minimization using adversarial learning as a deterministic posterior approximation technique using the KL divergence.
This is a similar procedure to the one considered in the main paper, but minimizes a different divergence and matches reconstructive statistics over marginal ones as performed with {\bf JSD}.

\subsubsection{Adversarial Inference With Tractable Likelihoods}
\label{kl_div_tractableLK}
The first learning rule is appropriate when we have explicitly stated models using the log-likelihood. Good-looking samples have been obtained in previous literature by blending adversarial losses and reconstruction losses and here we derive a principled explanation for some instances of them.

We assume $D(z,x) = \frac{p(z)}{q(z|x)+p(z)}$ and $(1-D(z,x)) = \frac{q(z|x)}{q(z|x)+p(z)}$.

\begin{equation}
\begin{split}
\text{log}p(x) &= \int \limits_{z} q(z|x) \text{log}p(x|z)dz - \int \limits_{z} q(z|x) \text{log} \frac{q(z|x)}{p(z)} dz + \text{KL}(q(z|x)||p(z|x))\\
\text{log}p(x) & \geq \int \limits_{z} q(z|x) \text{log}p(x|z)dz - \int \limits_{z} q(z|x) \text{log} \frac{q(z|x)}{p(z)} dz\\
&= \int \limits_{z} q(z|x) \text{log}p(x|z)dz - \text{KL}(q(z|x)||p(z)) \\
& = \int \limits_{z} q(z|x) \text{log}p(x|z)dz - \int \limits_{z} q(z|x) \text{log} \frac{1-D_{z}(z,x)}{D_{z}(z,x)} dz\\
& = \mathcal{L}_{rec}(x|\theta,\phi) - \int \limits_{z} q(z|x) \text{log} \frac{1-D_{z}(z,x)}{D_{z}(z,x)} dz\\
\end{split}
\end{equation}

We can easily draw samples for $p(z)$ and $q(z|x)$ from the prior and inference model, respectively, and can thus easily train a powerful classifier $D_{z}$ to perform the required discrimination.

This setting is particularly useful when combining adversarial training with tractable likelihoods and intractable posteriors and matches the model used for Adversarial Autoencoders~\citep{makhzani2015adversarial}.

\subsubsection{Adversarial Variational Inference With Intractable Likelihoods}
\label{adversarialKLZ}
For $q(x)$ being the true data distribution represented by samples of a dataset and $p(z)$ a prior, we assume $1-D_{z}(z,x) = \frac{p(z)}{q(z|x)+p(z)}$ and $(D_{z}(z,x)) = \frac{q(z|x)}{q(z|x)+p(z)}$. We furthermore similarly assume that $D_{x}(x,z) = \frac{q(x)}{q(x)+p(x|z)}$.
Then, we can sidestep the results from Section~\ref{kl_div_tractableLK} which require explicit evaluation of a reconstruction likelihood. We achieve this by minimizing the \text{KL}-divergence of $q(z,x)$ and $p(z,x)$ and performing adversarial inference on the resulting loss function.
This allows us to minimize reconstructive divergence terms without explicit likelihood evaluations.

\begin{equation}
\begin{split}
\text{KL}(q(x,z)||p(x,z)) &= \int \limits_{x,z} q(x,z) \text{log}\frac{q(x,z)}{p(x,z)} dxz\\
&=  \int \limits_{x} q(x) \int \limits_{z}  q(z|x) \text{log}\frac{q(x,z)}{p(x,z)} dz dx\\
&=  \int \limits_{x} q(x) \int \limits_{z}  q(z|x) \text{log}\frac{q(x)q(z|x)}{p(z)p(x|z)} dz dx\\
&=  \int \limits_{x} q(x) \int \limits_{z}  q(z|x) \Big[ \text{log}\frac{q(z|x)}{p(z)} +\text{log}\frac{q(x)}{p(x|z)} \Big] dz dx\\
&=  \int \limits_{x} q(x) \Big[ \text{KL}(q(z|x)||p(z))  + \int \limits_{z}  q(z|x)\text{log}\frac{q(x)}{p(x|z)} dz \Big]  dx\\
&=  \int \limits_{x} q(x) \int \limits_{z}  q(z|x) \Big[ \text{log}\frac{D_{z}(x,z)}{1-D_{z}(x,z)} +\text{log}\frac{D_{x}(x,z)}{1-D_{x}(x,z)} \Big] dz dx\\
\end{split}
\end{equation}

This framework reveals how a carefully chosen adversarial cost and an explicit likelihood represent related.
This is intuitively performed in various papers in previous literature~\citep{dosovitskiy2016generating,nguyen2016synthesizing} and explained formally here.

\subsubsection{Mixed Adversarial Variational Inference clarifies relation of adversarial learning and likelihood maximization}
\label{adversarialKLZmixed}
In Sections~\ref{kl_div_tractableLK} and~\ref{adversarialKLZ} we show how KL divergence minimization can lead to adversarial objective functions for tractable and intractable likelihoods. It is easy to see that the two objectives shown are related given for optimal discriminators and known likelihoods, since the regularizer involving the latent variable ($\text{KL}(q(z|x)||p(z))$) is the same.
The rest of the respective objective functions uses an explicit likelihood to score how near samples form the model are in the tractable case and an adversary that decides whether reconstructions are close enough to the original image in the latter case. This permits likelihood-free inference, which can be useful for undefined observation models.

When performing likelihood maximization we commonly use the whole dataset {\bf X} to maximize $\text{log}p(x)$. In this case, the related structure of the two objectives above is revealed as follows. We omit writing the adversaries for the regularizer since they have been clarified above.
When maximizing a likelihood, the evidence lower bound ({\bf ELBO}) yields the following result as shown above:
\begin{equation}
\text{log}p({\bf X}) \geq \sum _{x \in {\bf X}} \Big [ \mathcal{L}_{rec}(x|\theta,\phi) - \text{KL} (q(z|x)||p(z)) \Big].
\end{equation}

Optimizing this objective is equivalent to minimizing $\mathcal{L}_{\bf ELBO}(x|\phi, \theta, \xi)$ :
\begin{equation}
\mathcal{L}_{\bf ELBO}({\bf X}|\phi, \theta, \xi) = \sum _{x \in {\bf X}} \Big [ - \int \limits_{z} q(z|x) \text{log}p(x|z)dz+ \text{KL} (q(z|x)||p(z)) \Big].
\end{equation}

In the likelihood-free formulation, we are minimizing a slightly different divergence. However, the resulting loss bears a striking similarity:
\begin{equation}
\mathcal{L}_{\text{KL}} ({\bf X, Z}|\phi, \theta, \xi)=  \sum_{x \in {\bf X} } \Big[ - \int \limits_{z} q(z|x) \text{log}\frac{1-D_{x}(x,z)}{D_{x}(x,z)} dz+ \text{KL} (q(z|x)||p(z)) \Big] 
\end{equation}

This clarifies how adversarial training is related to maximum likelihood in a graphical model.

Furthermore, it yields potential insights into why we observe beneficial regularization effects when combining both approaches, since they correspond to the same criterion but are calculated in different ways.
Optimization-wise, it may confer benefits for the learning of the discriminator to blend its cost with an explicit likelihood or regularizer on the latent variable, if such an explicit parametric form is known, since the explicit likelihood acts as a variance reduction rechnique for the adversary.
Similarly, this can be chosen at any factor in a graph: applying the
trick of replacing ratios with adversaries can be used at will at every factor, since the objective is not affected.

As such, we have shown that for generative modeling it is still a separate task to determine a model from its explicit learning and inference algorithm. Additionally, the choice of divergence and overall learning procedure is unrelated to picking adversarial or likelihood-based learning.
Both stem from the same objective and should be used where appropriate to facilitate robust approximate inference in graphical models.
Adversarial learning can better cope with intractable distributions at the cost of potential saddle points during optimization while explicit likelihood-based learning is stable at the cost of complexity in the variational approximation it induces.

\subsection{Feature-based Message Passing}
An alternative representation stems from a feature view on density ratios. The introduction  of maximum mean discrepancy~\citep{gretton2012kernel} provides the theoretical underpinnings to understand any distribution as a point in an adequately complicated vector space and a two-sample test to depend on the statistics on the distances betweeen different distributions represented by points in that space. The basis of many divergences is the evaluation and minimization of expectations of ratios or, in the case of the {\bf JSD}, a softmax ratio between two distributions. In the context of MMD, this corresponds to minimizing distances in appropriate spaces between the approximate and the true distributions.

MMD-networks~\citep{dziugaite2015training} use this methodology as a means to learn generative models and our framework fits this as well.

\subsection{Divergence Minimization and Generation With Nonparametric Observation Models}
\label{sec:np_models}

Currently, sampling from $q(x|\tilde{\text{pa}}(x))$ is typically implemented using the reparametrization trick and generalizations thereof and takes the form:
\begin{equation}
q(x|\tilde{\text{pa}}(x)) = \int \limits_{\epsilon} p(\epsilon)g_{rt}( f_{pm}(\tilde{\text{pa}}(x)), \epsilon )d\epsilon.
\end{equation}
where $f_{pm}$ is a mapping (for instance a neural inference network) from an input to a parametric variational family.

We propose to free variational families from their parametric corsets and parametrize a more flexible variational family through a nonlinear function $ f_{vf}$. We directly sample from the approximate posterior by injecting the noise vectors as additional inputs into the nonlinear transformation of the parents, $x^{l} = f_{vf}( \tilde{\text{pa}}(x), \epsilon_{l})$.
A (not necessarily normalized) variational family is thus modeled by:
\begin{equation}
q^{*}(x|\tilde{\text{pa}}(x)) = \int \limits_{\epsilon} p(\epsilon) f_{vf}(\tilde{\text{pa}}(x),\epsilon) d \epsilon. 
\end{equation}
The subtle but powerful difference is that now the samples $x^{l}$ can represent an arbitrary distribution, constrained only by the capacity of the nonlinear function $f_{vf}$ and the dimensionality of the noise vector $\epsilon_{l}$. A Gaussian Process version was introduced idn the context of hierarchical variational families~\citep{tran2015variational}.
This trick also forms the basis of DISCO networks~\citep{bouchacourt2016disco} and was mentioned in the context of adversarial autoencoders ~\citep{makhzani2015adversarial}.
However, we re-introduce this trick as a general tool to represent rich variational families, which are a good fit with our flexible adversarial message passing framework, thereby generalizing from the specific cases mentioned ahead to a general approximate inference framework.
Specifically, previous variational inference techniques require a parametric form of the approximate posterior, such as obtained when using the reparametrization trick, in order to evaluate the divergence term needed to regularize learning. Within our framework, this divergence term is implicitly represented through samples which are scored within the adversarial framework, relieving the probabilistic modeler of the need to choose an explicit parametric form for approximate posterior families.
Together with other recent powerful advances in variational inference, such as the generalized reparametrization gradient~\citep{ruiz2016generalized} and a rejection sampling generalization~\citep{naesseth2016rejection} which both learn explicit transformations $h(g(\cdot))$ to represent complex parametric variational families, this enables practical use of complicated modeling assumptions which are not limited by tractability of the typically occuring ratios within many divergence terms. We also note the concurrently published work~\citep{operatorVI}, which focuses on a related idea irrespective of the link to adversarial inference, but gives deeper theoretical insights into the applicability of the same trick and provides further justification for our application thereof.
Finally, we note that the same approach can also be used to specify implicit observation noise models in generative models, such as done in generative neural samplers as introduced in the original GAN paper~\citep{goodfellow2014generative}. While this is not explicitly mentioned in~\citep{goodfellow2014generative}, it is plausible that generative adversarial networks can learn arbitrary noise models that may be hard to represent analytically and the typically high-dimensional inputs to the networks can be interpreted to factorize into noise contributions and actual latent variables.

\end{document}